\documentclass[letterpaper, 10 pt, conference]{ieeeconf}  

\usepackage{amssymb}
\usepackage{amsmath}
\usepackage{algorithm}
\usepackage{algpseudocode}
\usepackage{graphicx} 
\usepackage{multirow} 
\usepackage{booktabs}
\usepackage{tabularx}
\usepackage{caption}
\usepackage{ragged2e}
\usepackage{hyperref} 
\usepackage{lineno}
\usepackage{threeparttable} 
\usepackage{makecell} 
\usepackage{booktabs}
\usepackage{fancyhdr}

\IEEEoverridecommandlockouts

\title{\LARGE \bf
FABG : End-to-end Imitation Learning for Embodied Affective Human-Robot Interaction
}

\author{Yanghai Zhang$^{1}$, Changyi Liu$^{2}$, Keting Fu$^{1}$, Wenbin Zhou$^{2}$, Qingdu Li$^{1,\dagger}$, Jianwei Zhang$^{3}$%
\thanks{$^{1}$The authors are with Institute of Machine Intelligence, University of Shanghai for Science and Technology, Shanghai 200093, China. $^\dagger$ is the corresponding author, {\texttt{\small liqd@usst.edu.cn}}}%
\thanks{$^{2}$The authors are with Shanghai Droid Robot Co., Ltd. Shanghai 200433, China, {\texttt{\small lcy5656@outlook.com, wenbin@droidup.com}}}%
\thanks{$^{3}$The author is with the Department of Informatics, University of Hamburg, 20146 Hamburg, Germany, {\texttt{\small jianwei.zhang@uni-hamburg.de}}}%
}

\begin{document}

\maketitle
\thispagestyle{fancy}
\renewcommand{\headrulewidth}{0pt} 
\lhead{}
\lfoot{}
\cfoot{\small{This work has been submitted to the IEEE for possible publication. Copyright may be transferred without notice, after which this version may no longer be accessible.}}

\begin{abstract}

This paper proposes FABG (Facial Affective Behavior Generation), an end-to-end imitation learning system for human-robot interaction, designed to generate natural and fluid facial affective behaviors. In interaction, effectively obtaining high-quality demonstrations remains a challenge. In this work, we develop an immersive virtual reality (VR) demonstration system that allows operators to perceive stereoscopic environments. This system ensures that "the operator's visual perception matches the robot's sensory input" and "the operator's actions directly determine the robot's behaviors" - as if the operator replaces the robot in human interaction engagements. We propose a prediction-driven latency compensation strategy to reduce robotic reaction delays and enhance interaction fluency. FABG naturally acquires human interactive behaviors and subconscious motions driven by intuition, eliminating manual behavior scripting. We deploy FABG on a real-world 25 degree-of-freedom (DoF) humanoid robot, validating its effectiveness through four fundamental interaction tasks:  affective interaction,  dynamic tracking,  foveated attention, and gesture recognition, supported by data collection and policy training.

\end{abstract}

\section{INTRODUCTION}

Humanoid robots have been increasingly deployed in human-centric scenarios, including critical domains such as educational assistance, behavioral guidance, and emotional companionship \cite{chevalier2020examining1,alnajjar2020humanoid2}. Research shows that nonverbal behaviors - particularly postural adjustments, gestural dynamics, and facial microexpressions - play an essential role in the transmission of internal states during human interactions \cite{krauss1996nonverbal3,saunderson2019robots4}. Within this context, imitation learning (IL), as a prominent reinforcement learning paradigm, demonstrates significant advantages by enabling robots to acquire skills through expert behavior emulation \cite{zheng2022imitation5}. Existing studies have validated the effectiveness of IL in robotics, where robots achieve task execution ranging from simple manipulations to multi-step complex operations via human demonstration observation \cite{hua2021learning6,reggia2018humanoid7,schaal1999imitation8,zare2024survey9}. These capabilities establish IL as a promising approach for driving autonomous affective interactions in humanoid robots. However, this presents two core challenges: how to efficiently acquire high-quality facial expression and motion demonstration data, and how to design imitation learning strategies suitable for facial expression interaction scenarios.
\begin{figure}[h]
    \centering
    \includegraphics[width=\linewidth]{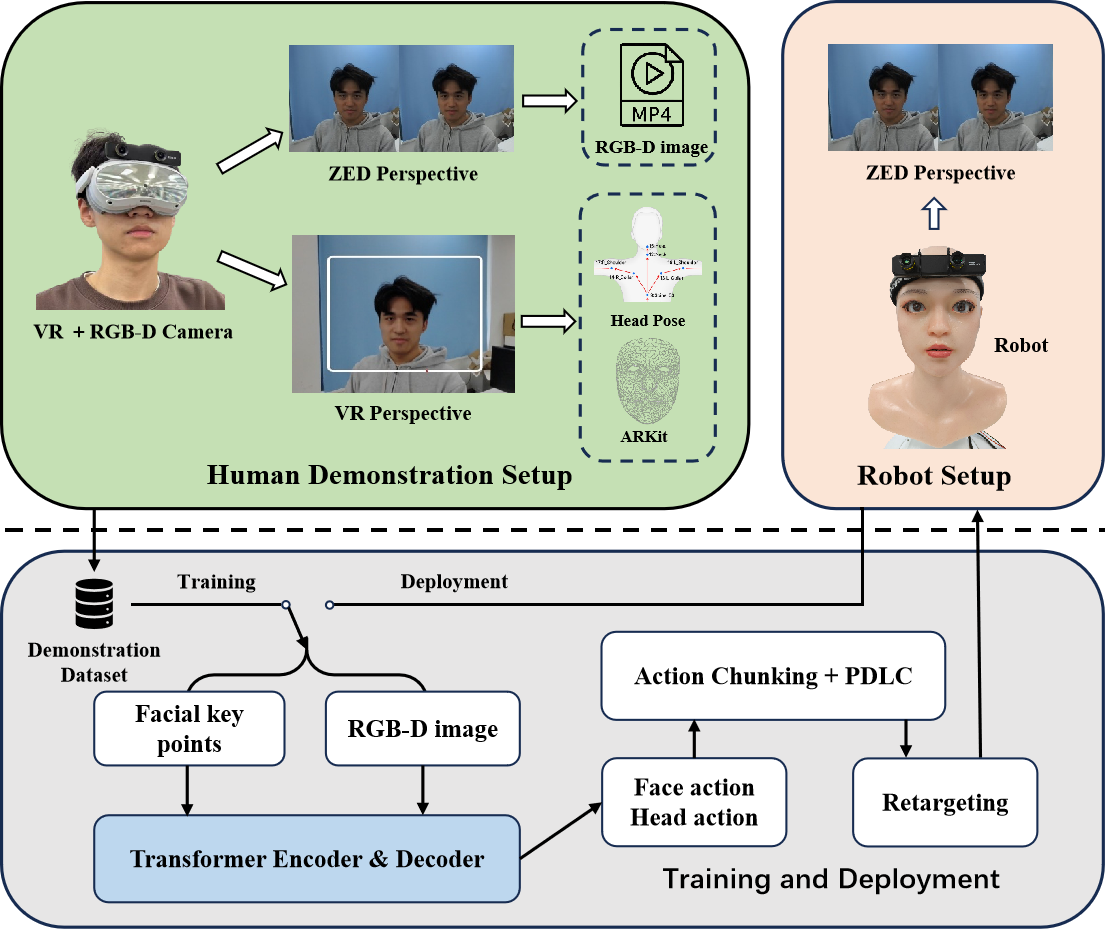}
    \caption{Human-to-robot behavior transfer. Top: VR captures facial landmarks/head pose (operator) and stereo RGB-D (ego-view). Bottom: We use the action chunking transformer to train an imitation policy for each task.}
    \label{fig:fig1}
\end{figure}

Current mainstream demonstration data acquisition methods are based on robotic teleoperation systems. Although existing studies have adopted interactive tools such as VR/AR controllers \cite{brohan2022rt10,lin2024learning11,duan2023ar212}, 3D spatial mice \cite{chi2023diffusion13,zhu2023viola14}, smartphones \cite{mandlekar2018roboturk15}, and haptic devices \cite{shaw2023videodex16,peternel2013learning17}, these solutions universally exhibit practical limitations: high-precision equipment incurs substantial costs, control interfaces demonstrate notable latency, and user interaction logic lacks intuitiveness. Regarding teleoperation perception, some systems directly utilize the operator's eyes to observe the robot's task space, resulting in the operator's view being frequently occluded by the robot's body, thereby hindering real-time evaluation of action execution outcomes.

As a crucial paradigm for robotic skill acquisition, imitation learning methods have achieved remarkable progress in operational task learning \cite{kimtransformer18,belkhale2023hydra19,xie2024decomposing20}. The recently proposed Action Chunking with Transformers (ACT) \cite{zhao2023learning21} enhances the performance of complex manipulation tasks through a task decomposition mechanism that segments long-horizon tasks into manageable temporal chunk sequences, significantly improving learning efficiency and execution coherence. However, ACT exhibits two critical limitations: (i) the original model relies solely on RGB inputs without leveraging depth information, and (ii) its Temporal Ensemble suffers from historical error accumulation in real-time interaction scenarios while disabling Temporal Ensemble leads to action discretization and execution stuttering.

To address these challenges, we propose FABG - an end-to-end system for embodied affective interaction - as shown in \autoref{fig:fig1}, with three key technical breakthroughs: First, an immersive VR teleoperation system. By designing a dual-modality data acquisition architecture that integrates ZED stereo cameras into VR devices, we synchronously capture the robot's first-person visual streams with the operator's facial landmarks/head pose data, enabling efficient acquisition of high-quality human demonstrations. Second, a depth-enhanced observation representation. We extend ACT's RGB input to RGB-D multimodal input, where the depth channel enhances spatial comprehension of human body language through dynamic distance awareness. Third, a prediction-driven latency compensation strategy that achieves triple optimization via prospective prediction: mitigating historical error propagation, eliminating mechanical jitter from discrete actions, and compensating multi-source system delays.

Our principal contributions are as follows:
\begin{enumerate}
    \item We introduce a novel affective interaction system for expressive robots, enabling the direct transfer of human facial interaction capabilities to robots, significantly enhancing motion naturalness and authenticity in HRI.
    \item Within this system, we deliver three technical innovations: an \textit{Immersive VR Demonstration System}, \textit{Depth-Augmented Observations}, and a \textit{Prediction-Driven Latency Compensation} strategy.
    \item We validate and conduct ablation studies of FABG on a physical humanoid robotic platform.
\end{enumerate}

\section{RELATED WORK}

\subsection{Demonstration Platform}

The core of imitation learning lies in extracting effective policies from expert demonstrations. While Behavior Cloning (BC)-based teleoperation methods \cite{pomerleau1991efficient22} have gained attention for their direct policy transferability, teleoperating physical robots for data collection presents significant challenges. At the perceptual level, existing systems predominantly adopt operator-first-person perspectives \cite{ fu2024learning23,wang2024dexcap24} or third-person viewpoints \cite{lin2024learning25,qin2023anyteleop26,sivakumar2022robotic27} for environmental observation. Such approaches are prone to perceptual deviations due to visual occlusions, leading to missing visual features in policy learning. Open-TeleVision \cite{cheng2024open28} innovatively eliminates occlusions via an immersive 3D active visual feedback system, yet its stereoscopic vision transmission suffers network latency, and the tightly-coupled motion control mechanisms may induce operator motion sickness. Furthermore, these systems' strong reliance on physical robots substantially limits demonstration scenario flexibility. UMI \cite{chi2024universal29} mounts GoPro cameras on handheld grippers, eliminating physical robot dependencies and providing a portable interface for robot teaching in natural environments, which inspires our system.

\subsection{Imitation Learning for Robotic Manipulation}

As a key technique in imitation learning, BC effectively accomplishes tasks by replicating user demonstrations \cite{pomerleau1988alvinn30}. However, BC suffers from compound error accumulation \cite{ ross2011reduction31,chang2021mitigating32}, where the propagation of temporal prediction deviations causes policies to drift from training distributions, leading to irrecoverable states \cite{ross2011reduction33,tu2022sample34}. To address this, DAgger \cite{ross2011reduction33} and its variants \cite{kelly2019hg35,menda2017bayesian36} mitigate error propagation by enabling additional policy interactions and expert correction mechanisms, while their requirement for continuous human intervention significantly increases user burden. ACT improves long-horizon manipulation task efficiency through temporal task decomposition, yet its chunked execution mode tends to create action discontinuities. Furthermore, employing temporal ensemble mechanisms in real-time interaction scenarios induces execution stalls due to the Temporal Ensemble of historical errors. We propose predicting action sequences at each timestep, dynamically selecting optimal execution frames via preset latency offsets to generate precise yet smooth motion trajectories, while compensating multi-source system delays for rapid response.

\section{METHOD}

FABG enables the direct transfer of human demonstrations in natural environments to deployable robotic policies. Our system overview is illustrated in \autoref{fig:fig1}, designed with three core objectives:

\textbf{Portability and Efficiency:} The Immersive VR-based Demonstration System can be deployed in any environment and initiates data collection with near-zero setup time.

\textbf{Effectiveness:} Collected data must encapsulate sufficient information to capture natural and complex human micro-expressions and subconscious motions. Imitation learning policies should enable robots to effectively acquire demonstrated behaviors, achieving fluid human-robot interaction while eliminating mechanical artifacts.

\textbf{Reproducibility:} Researchers and enthusiasts should be capable of deploying and utilizing FABG across diverse humanoid robotic platform.

The following sections detail how we realize these objectives through hardware, software, and policy design.

\subsection{Immersive VR Demonstration System} 

\begin{figure}[h]
    \centering
    \includegraphics[width=\linewidth]{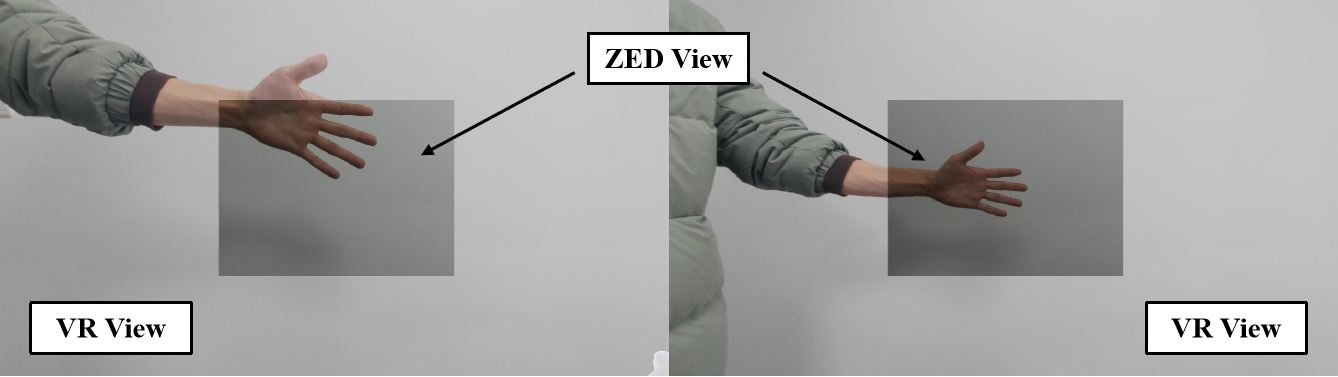}
    \caption{Alignment between PICO passthrough view and ZED camera feed for establishing observation space positioning in stereoscopic environments.}
    \label{fig:fig2}
    \vspace{-10pt}  
\end{figure}

Employing appropriate system for effective demonstration collection is critical to obtaining high-quality results. In this work, we present a virtual reality-based immersive system for collecting human user demonstrations. As shown in Fig. 1, human demonstrators utilize PICO 4 Pro VR headsets and ZED RGB-D cameras to perform interactive behaviors instead of physical robots. Compared to using two independent biomimetic cameras in robotic eyes, ZED RGB-D cameras provide stable and precise depth information. The VR system records the operator's 58 ARKit facial expression coefficients and head RPY (Roll, Pitch, Yaw) rotational movements in real-time, while the RGB-D camera synchronously captures 480×640 resolution first-person-view imagery. To equip operators with active first-person perception capabilities and align human-robot viewpoints, we developed a 3D environmental perception module in Unity, implementing stereoscopic passthrough functionality that enables operators to observe 3D environments consistent with the real world. As depicted in  \autoref{fig:fig2}, we calibrated the position and scaling of ZED camera feeds within the VR interface to achieve seamless stitching between 3D passthrough views and ZED camera fields of view. Spatial markers were then applied to the ZED video streams before disabling their display, thereby guiding operators into the correct visual zones.

\subsection{Depth-Augmented Observations}
To address spatial localization challenges in affective interaction scenarios, our solution improves upon existing methods that rely on binocular RGB images for indirect 3D geometry inference by synchronously processing multimodal sensory data. The system integrates dual 480×640×3 RGB streams and single-channel depth maps captured by stereo cameras at the input stage. The preprocessing stage implements a multi-stage strategy: Gaussian filtering with the following kernel is applied to depth maps for high-frequency noise suppression:
\begin{equation}
D_{\sigma}(x,y) = \frac{1}{2\pi\sigma^2} \exp\left( -\frac{x^2 + y^2}{2\sigma^2} \right)
\end{equation}

Here, $D_{\sigma}(x,y)$ denotes the Gaussian filter kernel where ${\sigma}$ controls blur intensity. Convolving this kernel with depth maps smooths fine details and removes noise. The feature extraction module employs a dual-path architecture: the RGB path extracts 384-dimensional semantic features via a pre-trained DinoV2 visual backbone, forming an $18\times24$ spatial feature tensor, while the depth path processes data through multi-layer CNNs to generate 128-dimensional geometric features with identical spatial resolution. Channel-wise concatenation constructs a $512\times18\times24$ fused feature matrix. This design preserves depth data's spatial representational strengths while integrating RGB's semantic comprehension, establishing a multimodal feature space for 3D interaction intent understanding. Experiments demonstrate this architecture effectively synergizes geometric structures and appearance semantics, providing a robust 3D perceptual foundation for natural human-robot interaction.

\subsection{Prediction-Driven Latency Compensation}

ACT generates precise and smooth motions through action chunking and temporal ensemble. However, this approach exhibits significant limitations in real-time interaction. When using action chunking alone, the robot generates a sequence of $k$ frames at time $t$ and executes them until time $t+k$. This discrete decision-making mechanism results in trajectory discontinuities at the junctions of adjacent action sequences, leading to periodic motion stuttering. The temporal ensemble avoids this issue by querying the policy at each time step. However, because it accumulates historical predictions, it inherently introduces response delays to immediate commands, potentially leading to error accumulation and amplification in dynamic environments due to noise interference. Moreover, multiple inherent system delays, such as the operator’s perception-action latency during data acquisition, computational delays in inference, and communication delays, further exacerbate response lag, ultimately impairing the real-time interaction experience between humans and robots.
\begin{figure}[h]
    \centering
    \includegraphics[width=0.6\linewidth]{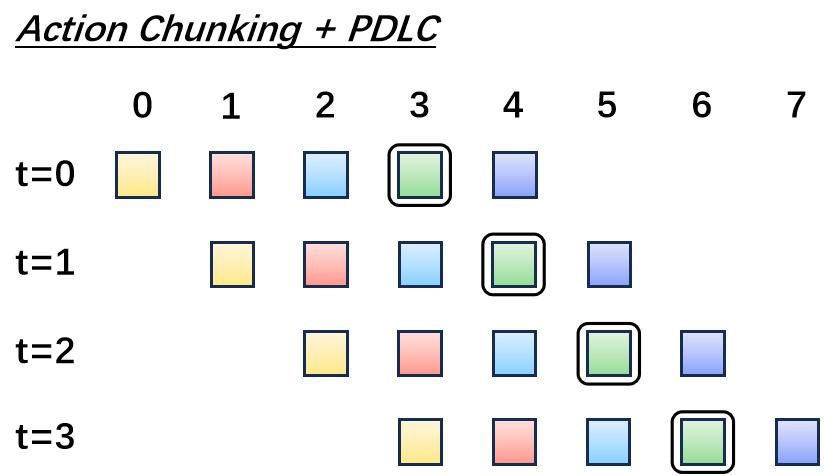}
    \caption{Our approach employs Action Chunking with PDLC, replacing conventional interleaved observation-execution cycles.}
    \label{fig:fig3}
\end{figure}

To address these issues, this study proposes Prediction-Driven Latency Compensation (PDLC). Our approach queries the policy at each time step to prevent periodic motion stuttering. Additionally, we design a dynamic truncation mechanism based on predefined time offsets. By quantitatively analyzing system delays, we determine the optimal action offset to achieve effective end-to-end latency compensation. Specifically, we fix the length of an action sequence to $k$, with $n$ representing the optimal action offset. At each time step, the robot receives an observation, generates an action sequence of length $k$, and executes the $(n+1)$-th action from the sequence, as illustrated in Figure 3. This process incurs no additional training cost or inference time. Experimental results demonstrate that PDLC is crucial for improving real-time human-robot interaction, as it significantly reduces response latency and generates precise and smooth interactive motions.

\begin{figure*}[h]
    \centering
    \includegraphics[width=0.8\linewidth]{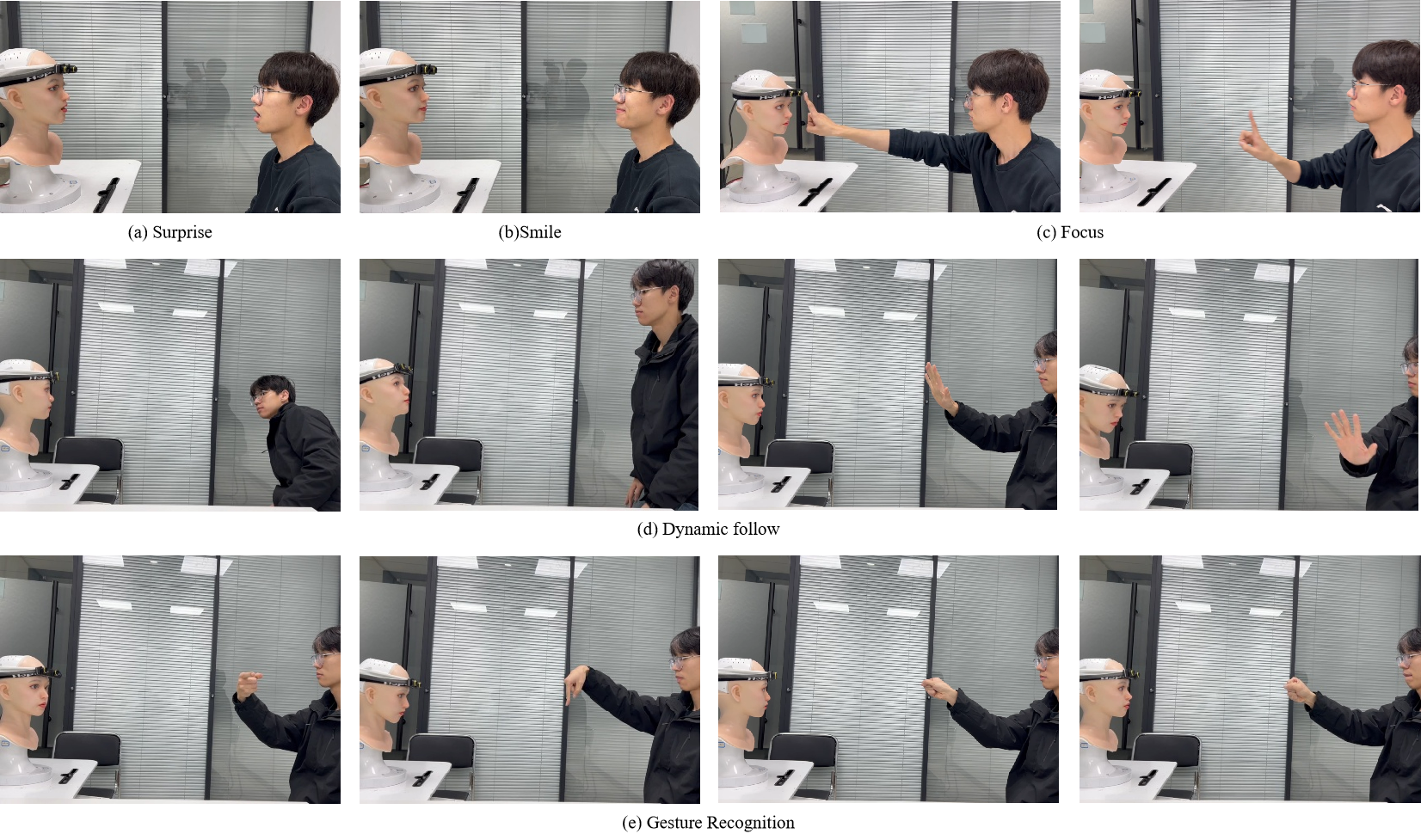}
    \caption{Human-Robot Interaction Scenarios in FABG}
    \label{fig:fig4}
\end{figure*}

\begin{table*}[htbp]
\centering
\caption{Task Completion Times.}
\begin{tabular}{cccccccc}
\toprule
\textbf{Input Modality} & \textbf{Execution Policy} & \multicolumn{2}{c}{\textbf{Affective Interaction}} & \multicolumn{2}{c}{\textbf{Dynamic Tracking}} & \textbf{Foveated Attention} & \textbf{Gesture Recognition} \\ \cmidrule(lr){3-4} \cmidrule(lr){5-6}
 &  & \textbf{Smile} & \textbf{Surprise} & \textbf{Face} & \textbf{Hand} &  &  \\ \midrule
RGB & w/o TE & --- & --- & 47.35 & 45.54 & 69.68 & 88.37 \\
RGB & w/ TE & 54.39 & 74.43 & 58.12 & 54.72 & 56.03 & 109.54 \\
RGB & PDLC(ours) & 10.54 & 12.03 & 14.61 & 10.75 & 17.71 & 18.17 \\
RGB-D(ours) & w/o TE & --- & --- & 43.58 & 39.56 & 41.42 & 79.05 \\
RGB-D(ours) & w/ TE & 45.62 & 53.60 & 54.39 & 52.28 & 43.84 & 98.94 \\
RGB-D(ours) & PDLC(ours) & \textbf{6.94} & \textbf{7.55} & \textbf{10.49} & \textbf{6.77} & \textbf{11.44} & \textbf{14.23} \\ \bottomrule
\end{tabular}
\begin{tablenotes}
    \item Results are averaged over five experimental trials. Time in seconds; '---' denotes task failure.
\end{tablenotes}
\label{tab:tab1}
\end{table*}

\section{EXPERIMENT}

\subsection{Hardware Platform}
We validate human-robot interaction behaviors using a self-designed bioinspired robotic head platform that simulates human-like facial expressions and head motions through a rigid-flexible hybrid structure and a multi-DoF actuation system. The core architecture comprises a 3D-modeled rigid framework integrated with elastic bioinspired skin. By converting expression parameters into 25-channel PWM signals to drive actuators, we achieve independent/coordinated control of key facial regions including eyelids, eyebrows, cheeks, and mouth. This platform establishes a controllable experimental interface for affective computing and behavioral mirroring research through engineering mapping of biomechanical characteristics.

\begin{figure*}[h]
	
	\begin{minipage}{0.32\linewidth}
		\vspace{3pt}
		\centerline{\includegraphics[width=\textwidth]{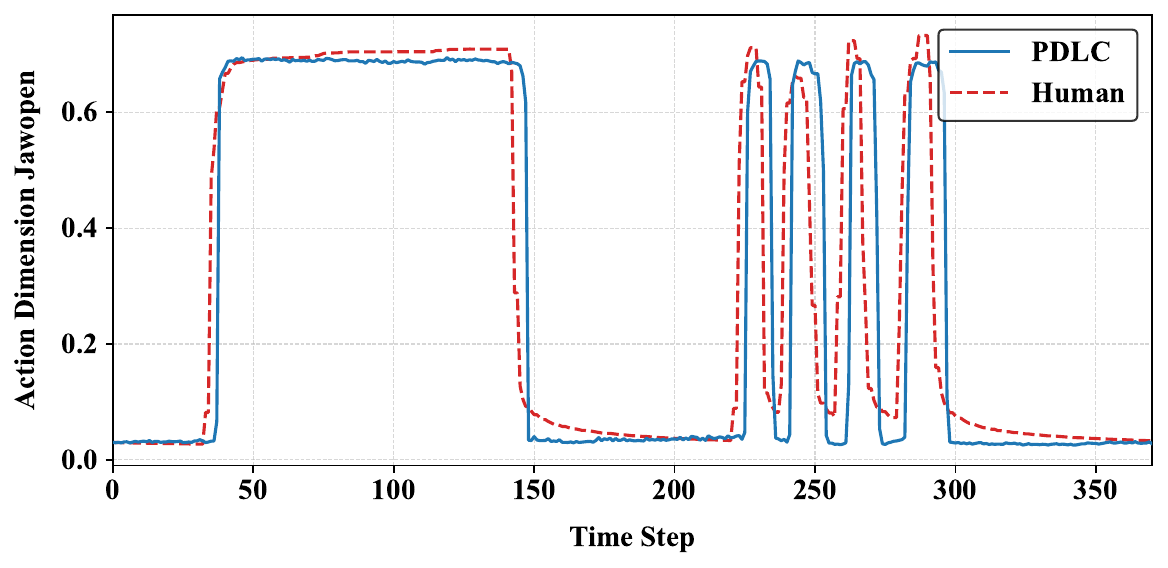}}
        
	\end{minipage}
	\begin{minipage}{0.32\linewidth}
		\vspace{3pt}
		\centerline{\includegraphics[width=\textwidth]{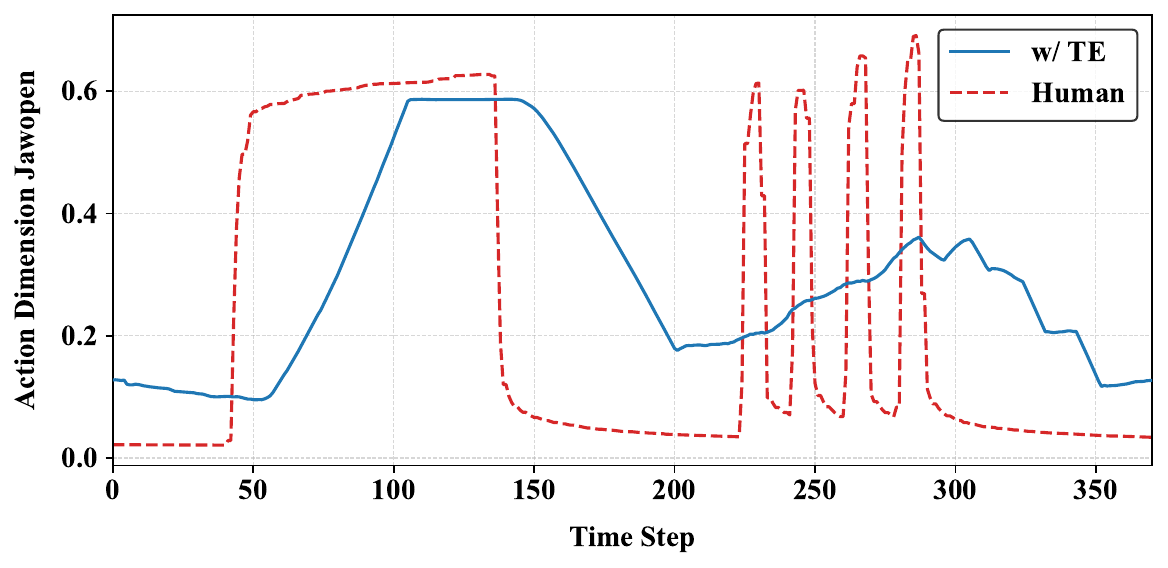}}
	 
	\end{minipage}
	\begin{minipage}{0.32\linewidth}
		\vspace{3pt}
		\centerline{\includegraphics[width=\textwidth]{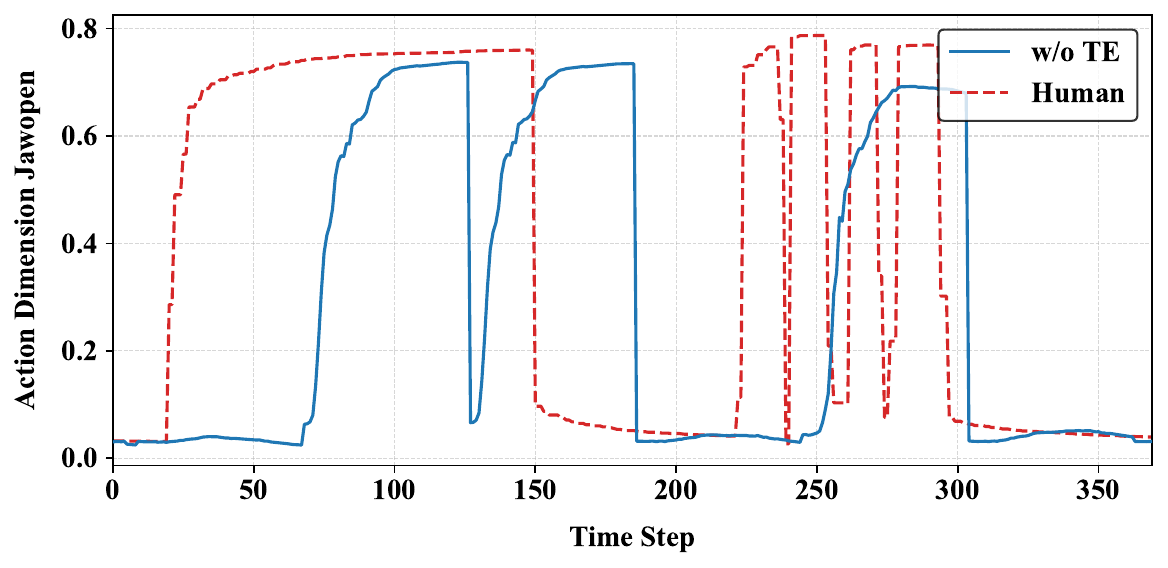}}
	\end{minipage}
 
	\caption{Comparative output curves of the JawOpen parameter generated by three strategies (PDLC, w/ TE, w/o TE) versus human demonstrations under identical model conditions. }
	\label{fig:fig5}
\end{figure*}

\subsection{Experimental Setup}
To validate the integrated performance of our method in real-time responsiveness, accuracy, and affective expressiveness, we design four representative human-robot interaction tasks to demonstrate the efficacy of the proposed data acquisition system and policy improvements, as shown in \autoref{fig:fig4}. The experimental configurations are as follows:

\textbf{Affective Interaction Task:} Requires real-time facial expression recognition and congruent response generation. The robot must reciprocate smiles with smiles and surprise expressions with corresponding reactions. When testers exhibit subtle variations in the intensity of expression, the robot synchronously adjusts its responses while maintaining mirroring during static expressions. The testers present 10 smiling expressions and 10 surprised expressions.

\textbf{Dynamic Tracking Task:} Evaluates head-eye coordination performance. The robot continuously tracks the tester's face until hand extension triggers gaze redirection to the hand. After the hand exits the robot's field-of-view (FoV), tracking reverts to facial focus. Each tester completes 10 trials for facial tracking and 10 for hand tracking.

\textbf{Foveated Attention Task:} Tests dynamic visual pursuit capabilities. The testers perform finger approach-retraction motions, during which the robot adaptively adjusts the interpupillary distance (IPD) to maintain gaze focus during proximity and resets IPD upon retraction. This task requires precise distance sensitivity for continuous finger tracking via dynamic IPD adaptation. Ten trials are conducted.

\textbf{Gesture Recognition Task:} Assesses dynamic command parsing. The testers maintain hand positions while randomly cycling through five gestures (up/down/left/right/fist). The robot must accurately recognize gestures to reorient toward specified directions or fixate on fists. 20 trials (4 per gesture) are executed.

For all four real-world tasks, we collect demonstrations via the immersive VR demonstration system. Task complexity dictates 2-5 second execution durations per trial by human operators, with randomized inter-trial intervals. We record 50 demonstrations per task, yielding total data set durations of 4-6 minutes.

\subsection{Experimental Results}

The input modalities cover both RGB and RGB-D configurations, while the execution strategies encompass three approaches: without temporal ensemble (w/o TE), temporal ensemble (w/ TE), and PDLC, resulting in six comparative experimental groups. All models are built upon the DinoV2 \cite{oquab2023dinov2} visual backbone and trained under unified computational conditions (NVIDIA RTX 4070Ti GPU) with identical hyperparameters: initial learning rate $5 \times 10^{-5}$, batch size 20, and 20,000 optimization iterations. To eliminate computational bias, all experiments execute inference at 30Hz. Validation is conducted on the facial robot platform described in Section IV-A, ensuring environmental consistency.

For all tasks, we evaluate mean task completion times across successful trials. To standardize timing, testers initiate subsequent interactions immediately after robot responses.

According to \autoref{tab:tab1}'s five-trial averaged results, our proposed PDLC strategy with RGB-D input demonstrates superior performance. In affective interaction tasks, models without inference strategies fail due to state instability, while RGB-D+PDLC achieves 6.94s (smile) and 7.55s (surprise) response times, showing 84.8 $\%$ and 85.9$\%$ improvements over RGB-D+TE. Depth sensing enhances micro-expression capture, reducing surprise response time by 37.2$\%$ compared to RGB+PDLC. For dynamic tracking, PDLC's motion trajectory prediction improves continuity, with RGB-D+PDLC completing hand tracking in 6.77s (37.0$\%$ faster than RGB+PDLC), confirming depth-enhanced 3D spatial localization. In foveated attention tasks, RGB-D+PDLC adjusts inter-pupillary distance in 11.44s (35.4$\%$ reduction vs. RGB+PDLC), benefiting from depth-based precise distance perception. Gesture recognition tasks show RGB-D+PDLC resolving commands in 14.23s (85.6$\%$ faster than TE), attributed to PDLC's temporal noise suppression. TE strategies increase average task durations by 21-25$\%$ due to historical interference, whereas PDLC reduces average time by 82.6$\%$ via dynamic compensation. Multi-group comparisons confirm that RGB-D and PDLC synergistically excel in spatial-sensitive tasks, providing robust solutions for real-time HRI systems.

\subsection{Ablation experiment}

To validate the effectiveness of PDLC in achieving low-latency response and high motion accuracy, this study designed a facial expression imitation experiment. Participants wearing motion capture devices executed standardized action sequences including sustained mouth-opening and rapid open-close motions, while their facial movements and the robot's response trajectories were synchronously recorded. Within the FABG system, a model with RGB-D input modality was employed to comparatively evaluate three configurations: the strategy without temporal reasoning (w/o TE), the temporal ensemble strategy (w/ TE), and the PDLC strategy.

The experimental results are shown in \autoref{fig:fig5}. The scheme without temporal reasoning suffers from action continuity loss due to discretized data processing. Although the robot can complete a single mouth-opening action within 0.55 seconds under this scheme, it exhibits a prolonged response time of 0.83 seconds and experiences action jitter due to an inability to sustain motion consistency. Temporal ensemble avoids action jitter by integrating historical information, but accumulating historical data leads to sluggish motion execution: while initiating responses within 0.38 seconds, it requires 0.86 seconds to complete the full mouth-opening process. In contrast, the PDLC strategy, through real-time optimization based on dynamic predictive compensation, maintains a low response latency of 0.116 seconds while compressing the full mouth-opening process to 0.05 seconds. Furthermore, the Dynamic Time Warping (DTW) distance between PDLC-generated motion sequences and human demonstrations is 8.82, representing reductions of 61.6$\%$ and 79.9$\%$ compared to the strategy without temporal ensemble (w/o TE: 22.95) and temporal ensemble strategy (w/ TE: 44.01), respectively. These metrics conclusively validate the PDLC strategy's superior performance in motion generation precision.

\section{CONCLUSIONS}

This study has successfully developed and validated FABG — an end-to-end embodied affective human-robot interaction system for real-world social scenarios. Addressing the limitations of conventional imitation learning strategies in affective interaction tasks, we propose an immersive VR-based natural teleoperation system that achieves high-quality human demonstration acquisition through streamlined user workflow design. The integration of depth-enhanced multimodal input mechanisms strengthens 3D spatial semantic parsing capabilities, complemented by a prediction-driven latency compensation strategy to optimize action temporal coordination. Multi-task interaction experiments on physical robotic platforms confirm system efficacy, with FABG demonstrating significant improvements in interaction accuracy and motion fluidity compared to ACT. Potential future work involves expanding FABG's multimodal integration, such as incorporating large language models (LLMs) for human-robot verbal interaction. This research provides technical foundations for humanoid robot applications in dynamic social contexts.

\bibliographystyle{IEEEtranS} 
\bibliography{references}

\end{document}